\documentclass[10pt,twocolumn,letterpaper]{article}

\usepackage{cvpr}              

\usepackage[accsupp]{axessibility}

\usepackage[hang,flushmargin]{footmisc}

\usepackage[most]{tcolorbox}

\definecolor{cvprblue}{rgb}{0.21,0.49,0.74}
\usepackage[pagebackref,breaklinks,colorlinks,allcolors=cvprblue]{hyperref}

\def\paperID{} 
\def\confName{CVPR}
\def\confYear{2026}

\title{PolyReal: A Benchmark for Real-World Polymer Science Workflows}

\author{
    Wanhao Liu$^{1,2*}$\quad
    Weida Wang$^{2,3*}$ \quad
    Jiaqing Xie$^{2,3}$ \quad
    Suorong Yang$^{7}$ \quad
    Jue Wang$^{1}$ \quad
    Benteng Chen$^{2,6}$ \\
    Guangtao Mei$^{1}$ \quad
    Zonglin Yang$^{8}$ \quad
    Shufei Zhang$^{2}$ \quad
    Yuchun Mo$^{4}$ \quad
    Lang Cheng$^{3}$ \\
    Jin Zeng$^{5}$ \quad
    Houqiang Li$^{1}$\quad
    Wanli Ouyang$^{2}$ \quad
    Yuqiang Li$^{2}$$^\dagger$ \\[2mm]
    $^1$University of Science and Technology of China \quad
    $^2$Shanghai Artificial Intelligence Laboratory \\
    $^3$Fudan University \quad
    $^4$Northwestern Polytechnical University \\
    $^5$Tongji University \quad
    $^6$The University of Hong Kong \\
    $^7$National University of Singapore \quad
    $^8$Nanyang Technological University \\[2mm]
    {\tt\small \{liuwanhao, wangweida,liyuqiang\}@pjlab.org.cn}
}
\begin{document}
\maketitle

\begin{abstract}
Multimodal Large Language Models (MLLMs) excel in general domains but struggle with complex, real-world science. We posit that polymer science, an interdisciplinary field spanning chemistry, physics, biology, and engineering, is an ideal high-stakes testbed due to its diverse multimodal data.
Yet, existing benchmarks related to polymer science largely overlook real-world workflows, limiting their practical utility and failing to systematically evaluate MLLMs across the full, practice-grounded lifecycle of experimentation.
We introduce \textbf{PolyReal}, a novel multimodal benchmark grounded in real-world scientific practices to evaluate MLLMs on the full lifecycle of polymer experimentation.
It covers five critical capabilities: (1) foundational knowledge application; (2) lab safety analysis; (3) experiment mechanism reasoning; (4) raw data extraction; and (5) performance \& application exploration.
Our evaluation of leading MLLMs on PolyReal reveals a capability imbalance. While models perform well on knowledge-intensive reasoning (e.g., Experiment Mechanism Reasoning), they drop sharply on practice-based tasks (e.g., Lab Safety Analysis and Raw Data Extraction). This exposes a severe gap between abstract scientific knowledge and its practical, context-dependent application, showing that these real-world tasks remain challenging for MLLMs. Thus, PolyReal helps address this evaluation gap and provides a practical benchmark for assessing AI systems in real-world scientific workflows.
\end{abstract}


\begingroup
\renewcommand{\thefootnote}{} 
\footnotetext{\noindent $^*$ These authors contributed equally to this work.}
\footnotetext{\noindent $^\dagger$ Corresponding author.}
\footnotetext{\noindent $^1$ All code and data can be found in \url{https://github.com/wanhaoliu/PolyReal}}
\endgroup    
\section{Introduction}
\label{sec:intro}

Multimodal Large Language Models (MLLMs), such as GPT-5~\cite{openai2025gpt5card} and Gemini-2.5 Series~\cite{comanici2025gemini}, have demonstrated remarkable capabilities in general domains, fluently reasoning across text, images, and data~\cite{hendrycks2020measuring, joshi2017triviaqa,wang2019superglue,zhou2025scientists}. This success has catalyzed immense enthusiasm for applying MLLMs to high-value and high-impact scientific fields~\cite{zheng2025large, swanson2025virtual,liu2025design,wang2025chem,liu2025moose}. However, their impressive performance in the ``generalist'' world obscures the formidable challenges they face when tackling the complex workflows of real-world scientific discovery.

We posit that polymer science, a cornerstone of modern materials, medicine, and engineering~\cite{miret2025enabling}, serves as the ideal and high-value testbed to evaluate MLLMs. The field is inherently interdisciplinary~\cite{lendvai2024rise}, a fact directly reflected in its diverse, expert-level sub-fields: for example, Photo-Functional Polymers draw heavily from chemistry and physics, Bio-Functional Polymers merge biology and chemistry, and Electro-Functional Polymers integrate principles of physics and engineering.

Given its importance, several benchmarks have begun to evaluate MLLMs on tasks related to chemistry and materials science~\cite{zhou2025scientists, laurent2024lab}. However, these efforts primarily focus on isolated, atomic tasks (such as simple chart question-answering) and are often limited to closed-form evaluation formats (e.g., MCQs). They fail to address the full scientific workflow —a dynamic, multi-step, and cross-modal process that defines real-world research. Crucially, they overlook the integrated capabilities required for a genuine lifecycle of experimentation: from applying foundational knowledge and assessing lab safety, to reasoning about mechanisms, parsing unstructured instrument data, and exploring final applications.

\begin{figure*}[h]
\centering
\resizebox{.85\linewidth}{!}{
\includegraphics[]{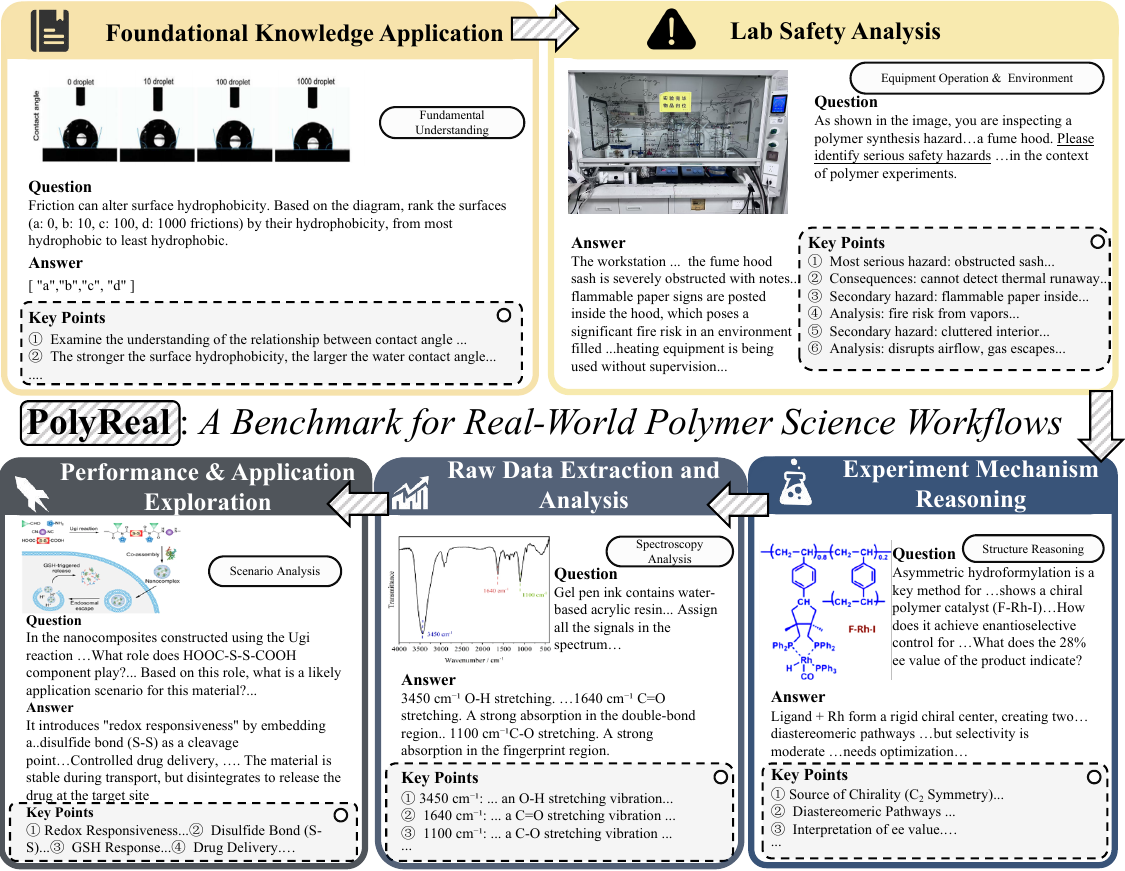}
}
\caption{Overview of the PolyReal benchmark and its five-stage workflow.} 
\label{fig:1} 
\end{figure*}

To bridge this critical gap, we introduce \textit{PolyReal}: a novel, multimodal benchmark grounded in real-world practices, designed to separately evaluate MLLMs on each critical stage of the polymer experimentation lifecycle.
\textit{PolyReal} covers five independent evaluation modules that define this workflow. An overview of these five modules is presented in Figure~\ref{fig:1}:
\begin{itemize}
    \item \textbf{Foundational Knowledge Application:} Evaluating the model's ability to apply core scientific principles to real-world scenarios.
    \item \textbf{Lab Safety Analysis:} Assessing visual scene understanding and risk identification in cluttered environments.
    \item \textbf{Experiment Mechanism Reasoning:} Testing causal and procedural reasoning based on reaction diagrams and chemical structures.
    \item \textbf{Raw Data Extraction and Analysis:} Measuring the ability to parse and digitize key information from diverse, unstructured instrument outputs, including visual charts (like Nuclear Magnetic Resonance (NMR) and Infrared (IR) spectroscopy) and raw data files (e.g., \texttt{.csv}).
    \item \textbf{Performance \& Application Exploration:} A high-level reasoning task that evaluates structure-property relationships and subsequently judges a material's potential suitability for a specific application.
\end{itemize}

Our contributions are threefold: (1) To the best of our knowledge, we introduce the first "full-workflow-coverage" benchmark grounded in real-world polymer science practice, composed of five independent, specialized modules that reflect the field's interdisciplinary nature. (2) We conduct a comprehensive evaluation of 15 leading MLLMs on \textit{PolyReal}, revealing a critical capability imbalance: models perform adequately on knowledge-intensive reasoning tasks but fail significantly on practice-based tasks.(3) Through in-depth qualitative analysis, we diagnose the root causes of these failures, such as a critical disconnect between abstract knowledge and its practical application, and factual hallucinations when parsing raw instrument data. This diagnosis pinpoints concrete optimization pathways for the next generation of AI systems designed to accelerate science.


\section{Related Work}

\label{Related Work}
\newcommand{\cmark}{\checkmark}
\newcommand{\xmark}{$\times$} 

Recent benchmarks have extended MLLM evaluation from general-purpose perception to scientific reasoning. General multimodal benchmarks, such as VQA~\cite{antol2015vqa,goyal2017making}, GQA~\cite{hudson2019gqa}, MMBench~\cite{liu2024mmbench}, LLaVA-Bench~\cite{liu2023visual}, and MMStar~\cite{chen2024we}, mainly evaluate visual understanding and general reasoning, but do not capture the domain knowledge and context-aware judgment needed in scientific research~\cite{hong2025worldsense,li2023comprehensive}.

Science-oriented benchmarks have moved evaluation toward expert domains. Early efforts, such as MMLU~\cite{hendrycks2009measuring}, SciEval~\cite{sun2024scieval}, AGIEval~\cite{zhong2024agieval}, and ChemLLMBench~\cite{guo2023can}, mainly focus on text-based scientific knowledge, while ScienceQA~\cite{auer2023sciqa} and TQA~\cite{kembhavi2017you} add multimodal inputs in largely educational settings. More recent domain-specific benchmarks include biomedical benchmarks built on medical images and clinical records~\cite{irvin2019chexpert,yan2017deeplesion,johnson2016mimic}, as well as chemistry and physics benchmarks such as OlympicArena~\cite{huang2024olympicarena}, OlympiadBench~\cite{he2024olympiadbench}, PhysUniBench~\cite{wang2025physunibench}, PhysX~\cite{shen2025phyx}, HiPhO~\cite{yu2025hipho}, JEEBench~\cite{arora2023have}, GPQA~\cite{rein2024gpqa}, CMPhysBench~\cite{wang2025cmphysbench}, PhyBench~\cite{qiu2025phybench}, ChemBench~\cite{mirza2024large}, and QCBench~\cite{xie2025qcbench}, which evaluate increasingly challenging scientific reasoning.

However, existing chemistry and materials benchmarks, such as MacBench~\cite{alampara2024macbench}, SFE~\cite{zhou2025scientists}, MatSciBench~\cite{zhang2025matscibench}, MSQA~\cite{cheung2025msqa}, and MatCha~\cite{lai2025can}, mainly emphasize isolated subtasks rather than the connected stages of real-world polymer research. In contrast, PolyReal focuses on workflow-oriented evaluation in polymer science, covering linked abilities from foundational knowledge and safety analysis to mechanism reasoning, data interpretation, and application exploration. Table~\ref{tab:benchmark_comparison} summarizes representative benchmarks.

\begin{table}[t]
    \centering
    \small
    \setlength{\tabcolsep}{4pt}
    \caption{Comparison with representative chemistry and materials benchmarks. MCQ: multiple-choice; Num.: numeric; EM: exact match; OQ: open question; T/F: true/false; Rank: ranking.}
    \label{tab:benchmark_comparison}
    \resizebox{\linewidth}{!}{%
    \begin{tabular}{l|l|c|c|c|l}
    \toprule
    \textbf{Benchmark} & \textbf{Domain} & \textbf{Size} & \textbf{Modality} & \textbf{Workflow} & \textbf{Type} \\
    \midrule
    MacBench    & Organic Chem. & 80  & Multi  & \xmark & MCQ, Num. \\
    SFE         & General Chem. & 285 & Multi  & \xmark & MCQ, EM, OQ \\
    MatSciBench & Polymer Chem. & 12  & Multi  & \xmark & Num., OQ \\
    MSQA        & Polymer Chem. & 257 & Single & \xmark & T/F, OQ \\
    MatCha      & Polymer Chem. & 98  & Multi  & \xmark & MCQ \\
    \midrule
    \textbf{PolyReal} & \textbf{Polymer Chem.} & \textbf{545} & \textbf{Multi} & \textbf{\cmark} & \textbf{OQ, Rank} \\
    \bottomrule
    \end{tabular}%
    }
\end{table}

\section{PolyReal Dataset and Tasks}

The \textit{PolyReal} benchmark is a multimodal evaluation platform focused on polymer science, comprising \textbf{545} high-quality question-answer pairs. Unlike many benchmarks that span multiple broad disciplines, \textit{PolyReal} is deeply focused on Polymer Science and its core innovation is the evaluation of the ``Full Polymer Research Workflow" paradigm within this complex, interdisciplinary field.

\subsection{Data Collection}
\label{sec:data_collection}

The construction of\textit{ PolyReal} is the result of extensive collaboration with domain experts in polymer science. Our data collection pipeline is designed to ensure authenticity, professional expertise, and breadth.
\paragraph{(1) Data Sourcing}
We first deconstructed the authentic research lifecycle of polymer science into five sequential, evaluable stages (our five modules), establishing the \textit{``full-workflow-coverage"} paradigm as the foundation of our benchmark. All data is sourced from real-world scientific scenarios. Experts curated high-value challenges from their authentic research (including ongoing experiments and published papers). Each task was designed to meet strict criteria: (1) It must reflect a meaningful, professional problem commonly encountered in research; (2) It must rely on predominantly authentic experimental data (e.g., visual charts, raw data files, lab scenes); (3) It must map clearly to one of the five workflow modules. 

\paragraph{(2) Content Breadth}
To ensure the benchmark's comprehensive coverage, the expert-constructed tasks systematically span multiple key sub-fields of polymer science. As detailed in our results (Figure~\ref{fig:statistic}b), this includes: Bio-Functional Polymers, Photo-Functional Polymers, Adsorption Polymers, Chemical-Functional Polymers, Electro-Functional Polymers, Smart Polymers, Nano-Functional Polymers, and Thermally Conductive Polymers.


\paragraph{(3) Question Format Curation}
To enable a robust evaluation of MLLM capabilities, our dataset covers diverse question formats. In addition to open-ended Q\&A and numerical extraction, we specifically designed \emph{34 Ranking Questions}. These ranking tasks demand a fine-grained, comparative understanding and effectively prevent models from achieving high scores through random guessing or shallow heuristics.


\paragraph{(4) Expert Authoring and Validation}
We instantiated the raw scientific data (e.g., spectra, \texttt{.csv} files, lab photos, mechanism diagrams) and research problems into VQA pairs. An expert team was responsible for authoring scientifically rigorous questions and their corresponding ground-truth answers. Subsequently, all instantiated VQA pairs underwent independent, expert cross-validation to ensure their scientific accuracy, clarity, and alignment with the objectives of their respective modules.

\subsection{Task}

\begin{figure*}[h]
\centering
\resizebox{1\linewidth}{!}{
\includegraphics[]{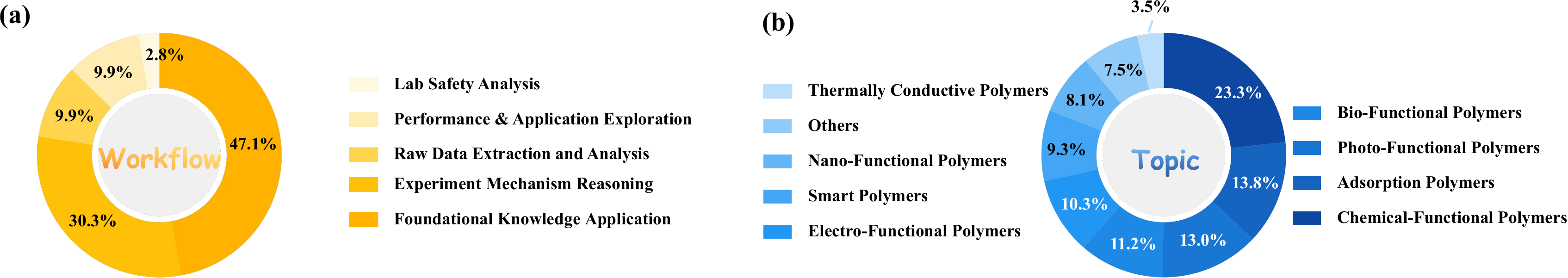}
}
\caption{Overview of the \textit{PolyReal} benchmark's composition, comprising 545 high-quality question-answer pairs: \textbf{(a)} Distribution of questions across the five core workflow modules covering the full research lifecycle; and \textbf{(b)} Comprehensive topic coverage, detailing the distribution across eight key sub-fields of polymer science.}
\label{fig:statistic} 
\end{figure*}

The \textit{PolyReal} benchmark is composed of five independent evaluation modules that sequentially mirror the full lifecycle of polymer science research. The distribution of the benchmark's 545 questions across these modules is visualized in Figure~\ref{fig:statistic}a.

The workflow begins with \textbf{Foundational Knowledge Application}, which evaluates the flexible application of core scientific principles to solve problems grounded in real experimental data and application scenarios, moving beyond simple text-based recall. This is followed by \textbf{Lab Safety Analysis}, assessing visual scene understanding and risk identification in cluttered, real-world lab photos. Next, \textbf{Experiment Mechanism Reasoning} tests deep causal and procedural reasoning on complex professional reaction diagrams, requiring models to infer reaction steps or determine the impact of changing conditions. Subsequently, \textbf{Raw Data Extraction and Analysis} assesses the parsing and digitization of key information from diverse, unstructured instrument outputs, including visual charts (e.g., NMR, IR spectra) and raw data files (\texttt{.csv}). Finally, \textbf{Performance \& Application Exploration} serves as a high-level capstone task that assesses a model's ability to evaluate structure-property relationships and judge a material's potential for a specific application.

\section{Experiments and Evaluations}
\label{sec:experiments}
\begin{figure}[h]
\centering
\resizebox{\columnwidth}{!}{
\includegraphics[]{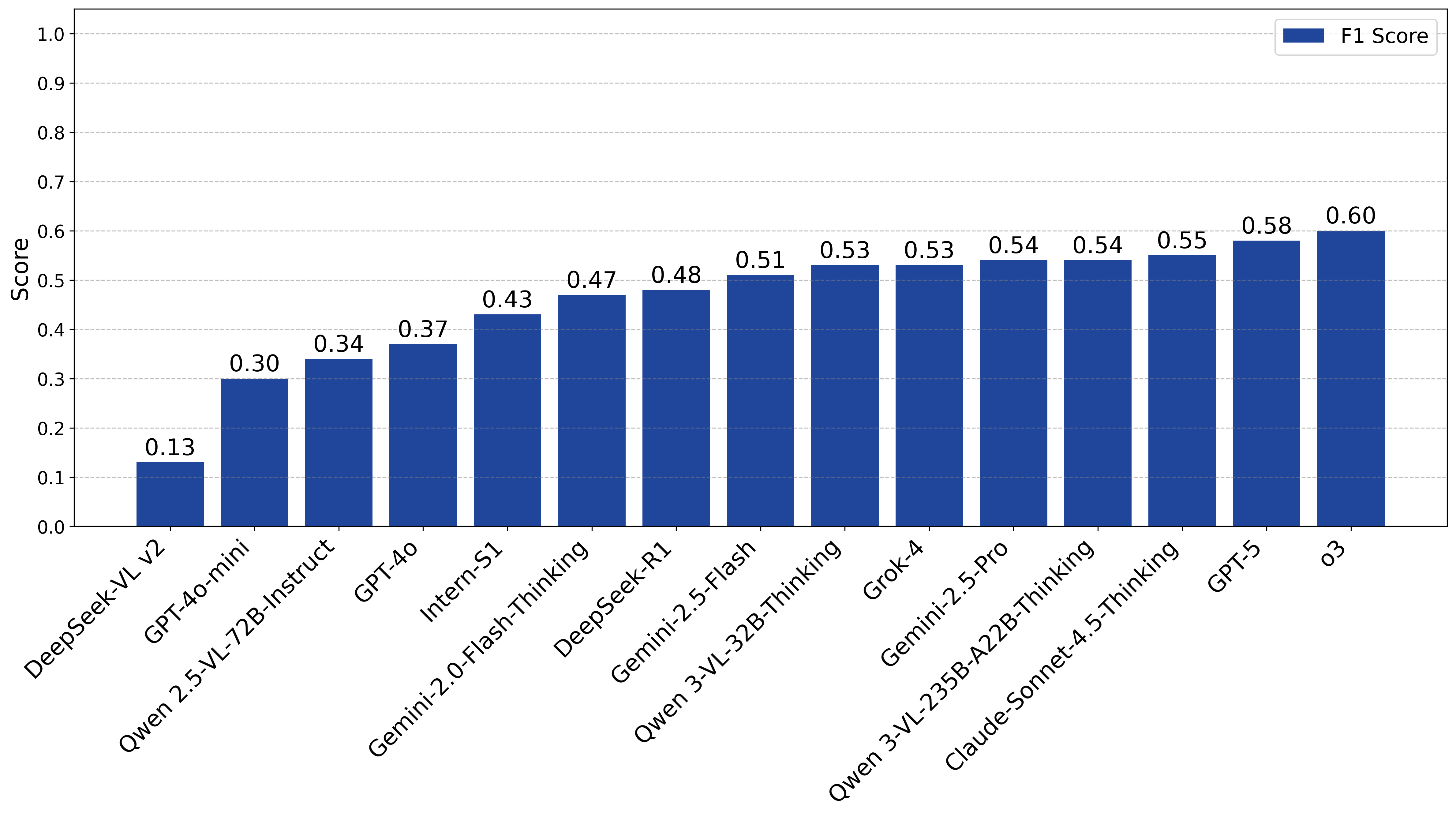}
}
\caption{Comparison of model performance across \textit{PolyReal}, evaluated by F1 score.} 
\label{fig:model_tasks_f1} 
\end{figure}

\subsection{Evaluation Setup}
\label{sec:eval_setup}

\paragraph*{Evaluated Models}
We evaluated a comprehensive suite of 15 leading Multimodal Large Language Models on the \textit{PolyReal} benchmark. As detailed in Table~\ref{tab:main_results}, these models are divided into two categories: (1) \textbf{Closed-Source Models}: O3~\cite{o3_o4-mini}, GPT-5~\cite{GPT-5}, GPT-4o~\cite{GPT-4o}, GPT-4o-mini~\cite{GPT-4o}, Grok-4~\cite{Grok-4}, Claude Sonnet 4.5~\cite{Claude-Sonnet-4.5}, Gemini 2.5 Pro~\cite{Gemini-2.5}, Gemini 2.5 Flash~\cite{Gemini-2.5}, and Gemini 2.0 Flash~\cite{Gemini-2.5}. (2) \textbf{Open-Source Models}: Qwen2.5-VL-72B-Instruct~\cite{bai2025qwen2}, Qwen3-VL-235B~\cite{Qwen3-VL}, Qwen3-VL-32B~\cite{Qwen3-VL}, DeepSeek-VL2~\cite{wu2024deepseek}, DeepSeek-R1~\cite{guo2025deepseek}, and Intern-S1~\cite{bai2025intern}. This selection enables a thorough comparison across diverse model architectures, scales, and training paradigms, covering both proprietary systems with state-of-the-art performance and publicly accessible models that support reproducible research.

\paragraph*{Metrics}
To comprehensively assess performance on \textit{PolyReal}, we adopt \textbf{Precision} ($P$), \textbf{Recall} ($R$), and the \textbf{F\textsubscript{1}-Score} ($F_1$) as our primary evaluation metrics. Given that \textit{PolyReal} includes a substantial number of open-ended question answering tasks, and ranking tasks, a simple binary accuracy metric is insufficient. To establish a rigorous, standard ground truth for these tasks, our domain experts meticulously delineated a set of ``Key Points" for each question, distilling the core scientific knowledge and experience required. The $P$/$R$/$F_1$ framework is then used to measure an answer's quality against these Key Points: \textbf{completeness} ($R$) measures whether the model identifies all correct points, and \textbf{correctness} ($P$) measures whether it avoids introducing erroneous information or irrelevant information, using both the Key Points and the Gold-Standard Answer as a reference. 

To validate the reliability of our automated evaluator (Gemini 2.5 Flash~\cite{Gemini-2.5}) against this framework, we conducted an inter-rater reliability study. We randomly sampled 80 non-ranking responses. Both our automated evaluator and human domain experts scored these responses by comparing them against both the ``Key Points" and the ``Gold-Standard Answer" to derive P and R scores. The two scoring methods showed exceptionally high agreement, achieving a Spearman's rank correlation ($\rho$) of 0.95 for the Recall ($R$) scores and 0.91 for the Precision ($P$) scores (see Table~\ref{tab:spearman}). This high correlation confirms that our automated metrics are a reliable proxy for expert judgment. The $F_1$-Score, as the harmonic mean, thus provides a robust single metric for evaluating complex scientific reasoning.


\begin{table}[h]
\centering
\caption{Spearman's ($\rho$) correlation between Human Expert and Automated Evaluator scores for P, R, and F1 metrics.\textsuperscript{*}}
\label{tab:spearman}
\begin{tabular}{lc}
\toprule
\textbf{Metric-Specific Score Correlation} & \textbf{Spearman's $\rho$} \\
\midrule
Recall ($R$) Scores (Completeness) & 0.95 \\
Precision ($P$) Scores (Correctness) & 0.91 \\
F1-Scores & 0.94\textsuperscript{*} \\
\bottomrule
\end{tabular}
\end{table}

\footnotetext{
\textsuperscript{*}F1-score correlation is calculated between the set of F1-scores derived from human P/R ratings and those derived from automated P/R ratings.
}

\vspace{-0.5cm}
\paragraph*{Implementation Details}
All experiments were conducted in a zero-shot setting, providing no in-context examples. This setup is designed to evaluate the ``out-of-the-box'' generalization capabilities of models without task-specific fine-tuning. For all model inference—whether via API or local execution—we set the decoding temperature to $0$ to minimize randomness and ensure the stability and reproducibility of our results.

\subsection{Main Results}
\label{sec:main_results}

We conducted a comprehensive evaluation of 15 leading MLLMs on the \textit{PolyReal} benchmark. The detailed Precision (P), Recall (R), and F1-Score (F1) results are presented in Table~\ref{tab:main_results}. Figure~\ref{fig:model_tasks_f1} provides a visual comparison of the overall F1 performance. Our experiments revealed several key findings.

\subsubsection*{ Observation 1: Overall Performance and Model Hierarchy}
As shown in Figure~\ref{fig:model_tasks_f1}, the overall model performance exhibits a clear hierarchy. Top-tier closed-source models, led by O3 (F1=0.60) and GPT-5 (F1=0.58), significantly outperform all other models on \textit{PolyReal}'s composite score. The second tier is formed by Claude-Sonnet-4.5-Thinking  (0.55) . In contrast, most open-source models (e.g., Qwen2.5-VL-72B-Instruct , F1=0.34) and some earlier models (e.g., DeepSeek-VL2, F1=0.13) show a distinct performance gap. This indicates that state-of-the-art closed-source MLLMs still hold a significant advantage in tackling the complex, interdisciplinary scientific tasks presented in \textit{PolyReal}.

\subsubsection*{ Observation 2: Efficacy of Knowledge Transfer vs. The Challenge of Real-World Practice}
Our in-depth analysis of the five modules (see Table~\ref{tab:main_results}) reveals a critical capability imbalance that directly reflects the composition of MLLMs training data.

First, models perform exceptionally well on the Foundational Knowledge Application (FKA) module (e.g., O3 achieves an F1 of 0.641). This is expected, as MLLMs training data (such as massive academic corpora) contains a large volume of this science-based knowledge, equipping them with strong foundational capabilities.

Encouragingly, we found that this strong foundational capability can be effectively transferred to high-level, knowledge-based reasoning tasks. This is demonstrated in the Experiment Mechanism Reasoning (EMR) and Performance \& Application Exploration (PAE) modules, which were the highest (or joint-highest) scoring modules for top models like O3 and GPT-5. This suggests that leading MLLMs do possess the potential to apply knowledge for complex reasoning in authentic scientific scenarios.

However, this capability transfer faces a severe challenge on practice-based tasks. The performance of all models drops significantly on the Lab Safety Analysis (LSA) and Raw Data Extraction (RDA) modules. This is particularly stark in the LSA task: among all 15 evaluated models, even the top performer, O3, only achieved an F1-Score of 0.412.

This by no means implies the LSA task is ``simpler"; rather, it highlights that these real-world tasks are exceptionally difficult for MLLMs. The LSA task requires the model to not only correctly `identify' chemical objects but also `understand' the environmental context and `synthesize' polymer science knowledge to perform `safety analysis'. This is likely because MLLM training data lacks a sufficient volume of multi-step, context-aware, practical data (such as complex lab scenes or unstructured instrument charts), leading to their capability gap in these modules.

\subsubsection*{Observation 3: High Recall vs. Low Precision (Analysis of Open-Ended Tasks)}
A common trend is observable in Table~\ref{tab:main_results}, especially in the open-ended EMR and PAE modules: nearly all high-performing models exhibit a much higher Recall (R) than Precision (P). For instance, GPT-5 achieves an R of 0.801 on EMR, while its P is only 0.543. This indicates that models tend to generate longer, more ``comprehensive" answers (high recall) in an attempt to cover all possible correct points, but do so at the cost of introducing a significant amount of irrelevant or erroneous information (low precision). This is a critical drawback where concise and accurate scientific answers are required.

\subsubsection*{Observation 4: Uneven Performance Across Sub-fields}
\begin{figure*}[h]
\centering


\resizebox{0.8\linewidth}{!}{
\includegraphics[]{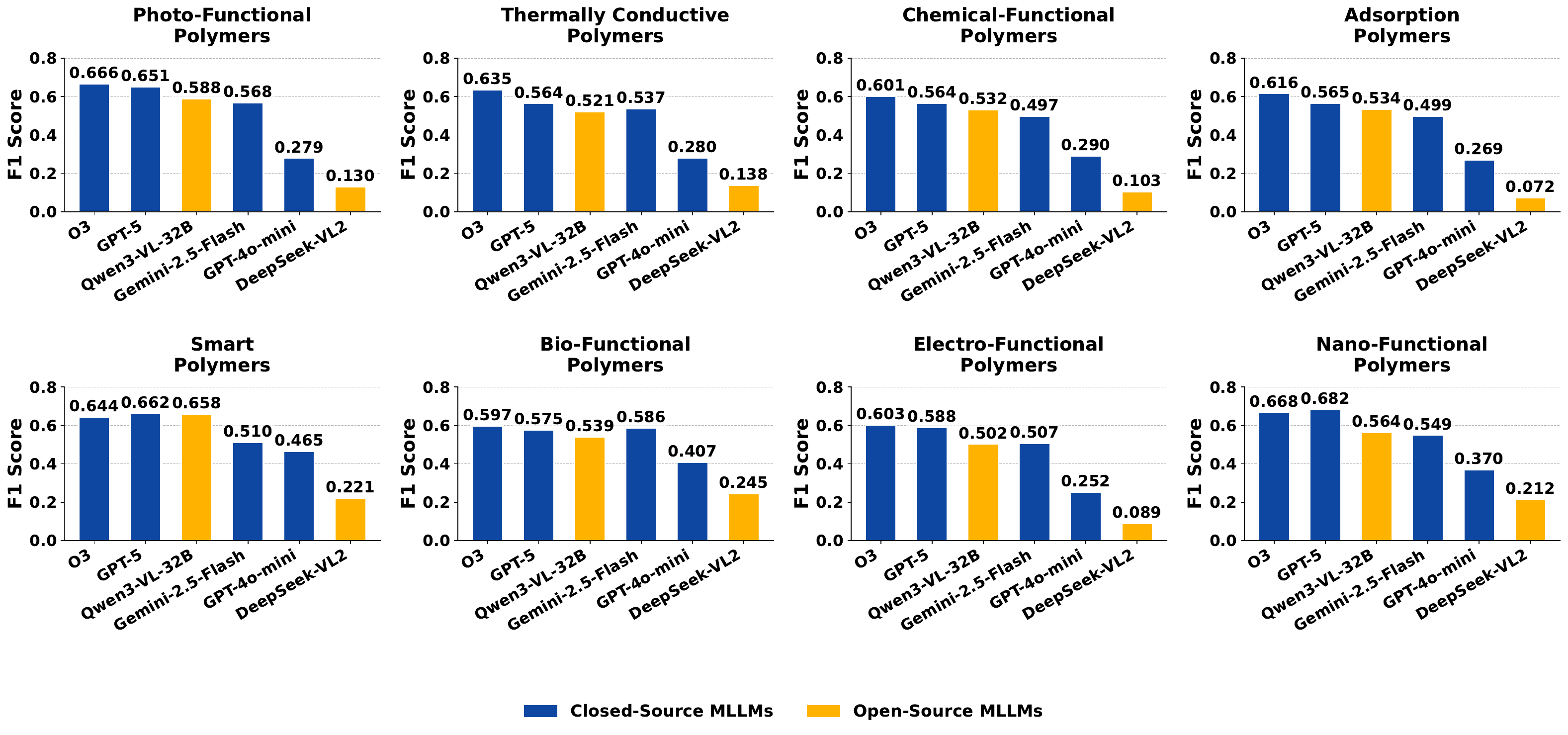}

}

\caption{Comparison of model performance across the eight key polymer science sub-fields in \textit{PolyReal}, evaluated by F1 score.}
\label{fig:polymer_class} 
\end{figure*}

Finally, as shown in Figure~\ref{fig:polymer_class}, the generalization of MLLMs capabilities across different polymer sub-fields is uneven. The analysis reveals that top models (like O3 and GPT-5) score highest on Nano-Functional Polymers and Smart Polymers. However, all models performed poorly in the Adsorption Polymers and Bio-Functional Polymers domains. This suggests that model performance is highly dependent not only on the task type (as seen in Observation 2) but also on the coverage of specific sub-domain knowledge in their training data.

\begin{figure*}[h]
\centering
\resizebox{0.8\linewidth}{!}{
\includegraphics[]{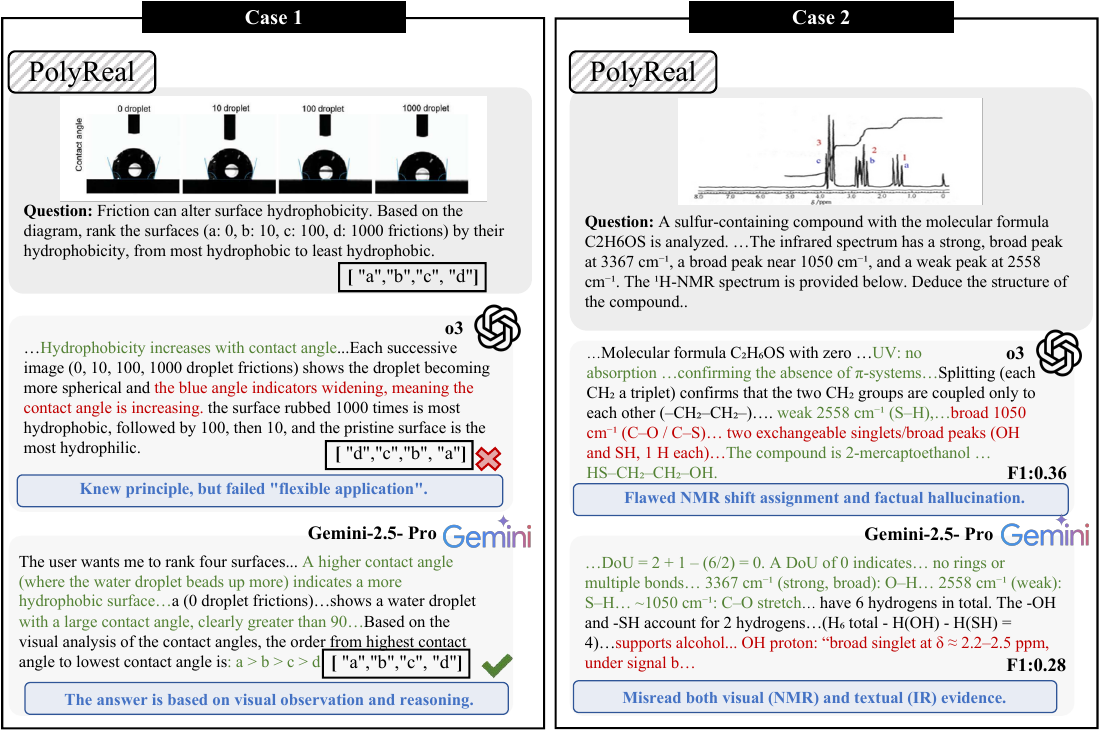}
}
\caption{Qualitative examples of MLLM failure modes: (Case 1) misapplying known principles and (Case 2) misinterpreting raw spectral data.}
\label{fig:case} 
\end{figure*}

\begin{table*}[t!]
\centering
\caption{Performance comparison of different models on various benchmarks. \textbf{P}, \textbf{R}, and \textbf{F1} represent Precision, Recall, and F1-score, respectively.\textbf{ LSA}: Lab Safety Analysis, \textbf{FKA}: Foundational Knowledge Application, \textbf{EMR}: Experiment Mechanism Reasoning, \textbf{RDA}: Raw Data Extraction and Analysis, \textbf{PAE}: Performance \& Application Exploration. The best results are in \textbf{bold} and the second best are \underline{underlined}.}
\label{tab:main_results}
\resizebox{\textwidth}{!}{%
\begin{tabular}{lrrrrrrrrrrrrrrrrrr}
\toprule
& \multicolumn{3}{c}{\textbf{FKA}} & \multicolumn{3}{c}{\textbf{LSA}} & \multicolumn{3}{c}{\textbf{EMR}} & \multicolumn{3}{c}{\textbf{RDA}} & \multicolumn{3}{c}{\textbf{PAE}} & \multicolumn{3}{c}{\textbf{Overall}} \\
\cmidrule(lr){2-4} \cmidrule(lr){5-7} \cmidrule(lr){8-10} \cmidrule(lr){11-13} \cmidrule(lr){14-16} \cmidrule(lr){17-19}
\textbf{Model} & \textbf{P} & \textbf{R} & \textbf{F1} & \textbf{P} & \textbf{R} & \textbf{F1} & \textbf{P} & \textbf{R} & \textbf{F1} & \textbf{P} & \textbf{R} & \textbf{F1} & \textbf{P} & \textbf{R} & \textbf{F1} & \textbf{P} & \textbf{R} & \textbf{F1} \\
\midrule
\multicolumn{19}{c}{\textit{\textbf{Closed-source Models}}} \\
\midrule
\textbf{O3} & 0.608 & 0.775 & \textbf{0.641} & \textbf{0.366} & 0.541 & \textbf{0.412} & 0.565 & 0.745 & \textbf{0.611} & 0.424 & 0.553 & 0.422 & 0.579 & \underline{0.733} & \underline{0.609} & 0.567 & 0.733 & \textbf{0.601}\\
\textbf{GPT-5} & 0.544 & \underline{0.799} & \underline{0.594} & 0.317 & \textbf{0.563} & \underline{0.373} & 0.543 & \underline{0.801} & \underline{0.608} & 0.418 & 0.523 & \underline{0.426} & 0.581 & \textbf{0.742} & \textbf{0.623} & 0.529 & \underline{0.760} & \underline{0.578} \\
\textbf{Claude-Sonnet-4.5-Thinking} & 0.519 & 0.752 & 0.580 & 0.282 & 0.411 & 0.317 & 0.539 & 0.700 & 0.578 & 0.354 & 0.489 & 0.356 & 0.526 & 0.663 & 0.539 & 0.503 & 0.692 & 0.546 \\
\textbf{Gemini-2.5-Pro} & 0.505 & 0.780 & 0.570 & 0.338 & 0.479 & 0.369 & 0.480 & 0.726 & 0.535 & 0.389 & \underline{0.592} & 0.408 & 0.521 & 0.677 & 0.555 & 0.483 & 0.726 & 0.536\\
\textbf{Grok-4} & \underline{0.633} & 0.614 & 0.570 & 0.321 & 0.267 & 0.224 & \underline{0.665} & 0.588 & 0.577 & 0.326 & 0.392 & 0.300 & \underline{0.598} & 0.549 & 0.540 & \underline{0.600} & 0.568 & 0.533 \\
\textbf{Gemini-2.5-Flash} & 0.439 & \textbf{0.816} & 0.530 & 0.294 & \underline{0.550} & 0.366 & 0.424 & \textbf{0.831} & 0.529 & 0.410 & \textbf{0.617} & \textbf{0.436} & 0.431 & 0.705 & 0.495 & 0.427 & \textbf{0.783} & 0.512 \\
\textbf{Gemini-2.0-Flash-Thinking} & \textbf{0.646} & 0.495 & 0.509 & \underline{0.342} & 0.260 & 0.267 & \textbf{0.667} & 0.437 & 0.466 & \textbf{0.448} & 0.338 & 0.332 & \textbf{0.656} & 0.440 & 0.468 & \textbf{0.625} & 0.450 & 0.468 \\
\textbf{GPT-4o} & 0.616 & 0.346 & 0.386 & 0.308 & 0.095 & 0.117 & 0.619 & 0.318 & 0.374 & \underline{0.429} & 0.311 & 0.323 & 0.592 & 0.323 & 0.370 & 0.587 & 0.323 & 0.367 \\
\textbf{GPT-4o-mini} & 0.550 & 0.312 & 0.346 & 0.266 & 0.063 & 0.078 & 0.489 & 0.222 & 0.258 & 0.302 & 0.212 & 0.211 & 0.572 & 0.309 & 0.351 & 0.501 & 0.268 & 0.299\\
\midrule
\multicolumn{19}{c}{\textit{\textbf{Open-source Models}}} \\
\midrule
\textbf{Qwen3-VL-235B-A22B-Thinking} & \underline{0.594} & 0.641 & \textbf{0.567} & \textbf{0.369} & \textbf{0.406} & \textbf{0.347} & \underline{0.585} & \textbf{0.655} & \textbf{0.578} & 0.368 & \underline{0.484} & \underline{0.355} & 0.549 & \underline{0.553} & \underline{0.518} & \underline{0.558} & \textbf{0.615} & \textbf{0.538}\\
\textbf{Qwen3-VL-32B-Thinking} & 0.544 & \underline{0.658} & \underline{0.554} & \underline{0.334} & 0.237 & 0.230 & 0.544 & \underline{0.619} & \underline{0.547} & \textbf{0.392} & \textbf{0.505} & \textbf{0.390} & 0.561 & \textbf{0.577} & \textbf{0.538} & 0.525 & \underline{0.612} & \underline{0.525}\\
\textbf{DeepSeek-R1} & 0.517 & \textbf{0.694} & 0.551 & 0.197 & 0.152 & 0.136 & 0.449 & 0.607 & 0.480 & 0.335 & 0.377 & 0.329 & 0.483 & 0.476 & 0.427 & 0.466 & 0.600 & 0.484\\
\textbf{Intern-S1} & 0.563 & 0.539 & 0.468 & 0.311 & \underline{0.264} & \underline{0.249} & \textbf{0.589} & 0.416 & 0.403 & \underline{0.389} & 0.368 & 0.304 & \underline{0.580} & 0.502 & 0.476 & 0.548 & 0.473 & 0.427 \\
\textbf{Qwen2.5-VL-72B-Instruct} & \textbf{0.607} & 0.321 & 0.370 & 0.298 & 0.107 & 0.144 & 0.567 & 0.264 & 0.312 & 0.323 & 0.245 & 0.245 & \textbf{0.612} & 0.373 & 0.402 & \textbf{0.559} & 0.295 & 0.337\\
\textbf{DeepSeek-VL2} & 0.373 & 0.129 & 0.158 & 0.310 & 0.024 & 0.036 & 0.350 & 0.080 & 0.101 & 0.216 & 0.124 & 0.138 & 0.388 & 0.102 & 0.122 & 0.350 & 0.108 & 0.132\\
\bottomrule
\end{tabular}%
}
\end{table*}
\subsection{Analysis}
\label{sec:case_studies}
To qualitatively investigate the ``capability imbalance" observed in our quantitative results (Section 4.2), we conducted an in-depth qualitative analysis of low-scoring task cases to analyze the limitations of model capabilities.

\subsubsection*{Analysis 1: Disconnect Between Knowledge and Real-World Application}
\label{sec:case_study_fka}

Our first key finding from this analysis is a severe disconnect between the ``knowledge" MLLMs possess and their ``real-world application" of it. By analyzing model responses (specifically, the `llm think' process), we found that models can often correctly ``recite" a scientific principle but fail when required to "flexibly apply" that principle to authentic, non-standard scientific data.

The FKA ranking task shown in Figure~\ref{fig:case} (Case 1) perfectly exposes this. The task requires the model to rank surfaces by hydrophobicity (from strongest to weakest) based on a series of water contact angle images. The correct scientific principle is that stronger hydrophobicity corresponds to a larger contact angle. The Gemini model arrived at the correct answer by correctly applying the scientific principle through a valid ``visual observation-reasoning'' chain. In stark contrast, the top-performing models, O3 and GPT-5, both produced the exact opposite, incorrect answer.

The failure mode of O3 is particularly illuminating. Its thought process (`llm think') shows it knew the correct scientific principle ("Hydrophobicity increases with contact angle"). Its failure, however, occurred at the application stage. The O3 model's reasoning explicitly states: "...the blue angle indicators widening, meaning the contact angle is increasing".

This is a Perceptual Error that is in direct contradiction to the visual evidence. The O3 model's understanding of "contact angle" is superficial; it failed to generalize the abstract concept to its concrete representation in a real-world scientific diagram (i.e., the instrument image), leading to its failure in practical judgment. This case strongly demonstrates that MLLMs suffer from a critical flaw in "flexible application," revealing a disconnect between knowledge and practice, even in high-scoring models.

\subsubsection*{Analysis 2: Failure in Reasoning Chains and Factual Hallucination under Data Scarcity}
\label{sec:case_study_rda}



Our second key finding is that models' reasoning chains are prone to failure when faced with complex, unstructured instrument data, often leading them to ``hallucinate'' non-existent evidence to rationalize their conclusions, consistent with our prior work~\citep{wang2025chem,li2025mol}. This is exemplified by the RDA task in Figure~\ref{fig:case} (Case 2), where models failed to deduce a structure by combining IR (text prompt) and NMR (image) data. The O3 model's failure was a Factual Hallucination (e.g., identifying a non-existent ``CH3 a triplet''). In contrast, the Gemini model's failure was a Perceptual and Attribution Error (e.g., misreading clear multiplets as a ``broad singlet''). This difficulty with raw instrument data is systematic: our analysis shows models similarly fail to correctly parse key features from polymer mechanical test \texttt{.csv} files, Thermogravimetric Analysis (TGA) charts, and X-ray Diffraction (XRD) patterns.

Interestingly, models performed slightly better when analyzing material morphology images, such as those from Scanning Electron Microscopy (SEM). This stark contrast, between partial success on visual morphology images (like SEM) and general failure on abstract spectral/chart data (like NMR, XRD, and TGA), re-confirms our conclusion. MLLMs possess a very limited ability to parse highly abstract ``real instrument data'' that requires specific domain knowledge (like spectroscopy or crystallography). This misaligned analysis is a critical bottleneck that must be addressed before MLLMs can accelerate scientific discovery.




\subsubsection*{Analysis 3: Limited Understanding of Unique Macromolecular Properties}
\label{sec:case_study_macro}

Our third key finding is that MLLMs generally lack a true understanding of core Macromolecular Science concepts, tending to "downgrade" them by applying flawed "small-molecule chemistry" logic.

To probe whether models understand the structural characteristics of macromolecules, \textit{PolyReal} intentionally includes a series of tasks comparing polymer reagents to their small-molecule counterparts. A clear pattern of failure emerged in these tasks.

For example, in a task concerning the design of high-performance quinone-based redox resins, models were asked to analyze the key to their high performance. From an expert's perspective, the core concept is the ``Site Isolation Effect": the polymer backbone effectively separates the active groups (quinone/hydroquinone) in space, thereby preventing side reactions (like dimerization) that are common in small-molecule solutions.

Our analysis shows that while models like O3 and GPT-5 could correctly identify the quinone as the active center (which is small-molecule knowledge), they almost entirely failed to mention the critical properties of the polymer chain itself. They did not reference the "site isolation" concept, unique to polymer chemistry, and instead analyzed the polymer reagent as if it were a simple small molecule.

This failure is significant. It indicates that LMM knowledge bases may be saturated with "small-molecule" facts but have not yet "emerged" a deep understanding of macromolecular properties (such as chain entanglement, steric hindrance, or site isolation). \textit{PolyReal} successfully reveals this critical gap in MLLMs' domain-specific scientific knowledge.

\subsection{Ablation}

\label{sec:ablation}



\begin{table}[h] \scriptsize
\centering
\caption{Ablation Study on the Importance of Modality. Performance comparison between the full multimodal input ("With Image") and the "Text-Only" (w/o Image) version.}
\label{tab:ablation_modality}
\begin{tabular}{lrrr}
\toprule
\textbf{Method} & \textbf{Precision ($P$)} & \textbf{Recall ($R$)} & \textbf{F1-Score ($F_1$)} \\
\midrule
\textbf{Gemini-2.5-pro} (w Image) & 0.41 & 0.79 & 0.51 \\
\quad w/o Image (Text-Only) & 0.36 & 0.71 & 0.45 \\
\midrule
\textbf{o3} (w Image) & 0.56 & 0.74 & 0.60 \\
\quad w/o Image (Text-Only) & 0.53 & 0.70 & 0.56 \\
\bottomrule
\end{tabular}
\end{table}

\begin{table}[h] \scriptsize
\centering
\caption{Ablation Study on the Role of External Knowledge.}
\label{tab:ablation_tool_use}
\begin{tabular}{lrrr}
\toprule
\textbf{Method} & \textbf{Precision ($P$)} & \textbf{Recall ($R$)} & \textbf{F1-Score ($F_1$)} \\
\midrule
\textbf{GPT-4o} (Closed-Book) & 0.58 & 0.30 & 0.35 \\
\quad w/ Search Tool & 0.56 & 0.38 & 0.41 \\
\midrule
\textbf{GPT-4o-mini} (Closed-Book) & 0.50 & 0.25 & 0.29 \\
\quad w/ Search Tool & 0.48 & 0.32 & 0.34 \\
\bottomrule
\end{tabular}
\end{table}

\subsubsection*{The Importance of Modality}
To quantify the contribution of the visual modality in \textit{PolyReal} and validate the role of multimodal information, we conducted an ablation study. We created a strict ``text-only'' version of the benchmark (labeled "w/o Image"), a setup in which we removed the visual inputs from the Foundational Knowledge Application (FKA), Experiment Mechanism Reasoning (EMR), and Performance \& Application Exploration (PAE) tasks. 

As shown in Table~\ref{tab:ablation_modality}, the results show a clear and significant drop in performance when visual data is removed. The F1-Score for Gemini-2.5-pro decreased from 0.51 to 0.45 (a drop of 0.06), and the F1-Score for the top-performing O3 dropped from 0.60 to 0.56 (a drop of 0.04). This finding provides strong evidence that PolyReal is a genuine multimodal benchmark; MLLMs cannot ``guess'' the answer from text prompts alone but must actually perceive and parse the visual information. It also reveals an important trend: the comparatively weaker model suffered a larger performance loss than the stronger base model (O3), suggesting that stronger models, though still impacted, are slightly less affected by the removal of visual data due to their more robust internal knowledge.

\subsubsection*{The Role of External Knowledge}
\label{sec:ablation_tool_use}

\textbf{Setup.}
In real-world scientific practice, scientists actively consult literature or search engines when facing knowledge bottlenecks. To investigate the capability differences of MLLMs in ``closed-book'' (relying only on internal knowledge) versus ``open-book'' (with access to external knowledge) scenarios, we conducted a second ablation study. We provided GPT-4o and GPT-4o-mini with access to a Search Tool (labeled "w/ Search Tool"). We focused on evaluating the tool's impact on performance in the knowledge-intensive modules: \textbf{FKA}, \textbf{EMR}, and \textbf{PAE}.

\textbf{Results and Analysis.}
We observed that after gaining access to the search tool, the models' performance on these knowledge-intensive modules improved significantly, as shown in Table~\ref{tab:ablation_tool_use}. Specifically, the F1-Score for GPT-4o \textbf{increased from 0.35 to 0.41} (a gain of \textbf{+0.06}), and the F1-Score for GPT-4o-mini \textbf{rose from 0.29 to 0.34} (a gain of \textbf{+0.05}). This confirms that MLLMs can effectively leverage external knowledge tools to partially bridge the knowledge gap when tackling complex, domain-specific scientific tasks.
\section{Conclusion}
\label{sec:conclusion}

In this paper, we introduced \textit{PolyReal}, the multimodal benchmark designed to systematically evaluate MLLMs on the \textit{real-world polymer science workflow}. Our benchmark fills a critical gap in existing scientific evaluation by providing five independent modules that simulate the research lifecycle. Our comprehensive evaluation of 15 leading MLLMs exposed a critical capability imbalance: models perform adequately on knowledge-based reasoning tasks but exhibit severe deficiencies on practice-based tasks. When faced with unstructured instrument data, their reasoning chains often fail, leading to \textit{factual hallucinations}, and demonstrating a limited understanding of core macromolecular concepts. We also confirm that  visual information is essential. \textit{PolyReal} serves not only as a clear diagnostic for the bottlenecks in current MLLMs but also paves the way for the next generation of AI Agents capable of accelerating scientific discovery.




\section*{Acknowledgments}
This work was supported by the New Generation Artificial Intelligence - National Science and Technology Major Project (No. 2025ZD0121802) and Intern-Discovery. Part of this work was done during the internships of the co-first authors at Shanghai Artificial Intelligence Laboratory.

{
    \small
    \bibliographystyle{ieeenat_fullname}
    \bibliography{main}
}
\newpage
\clearpage
\setcounter{page}{1}
\maketitlesupplementary


\section{Difficulty-Graded Evaluation Results}
To provide a more granular understanding of MLLM capabilities, we stratified the PolyReal dataset into three difficulty levels: Easy (280 samples), Medium (165 samples), and Hard (100 samples). This stratification allows us to distinguish between superficial perception capabilities and deep, expert-level scientific reasoning.
\subsection{Difficulty Grading Criteria}

To ensure a fine-grained assessment of MLLM capabilities, our domain expert team manually categorized the 545 samples into three difficulty tiers, including Easy, Medium, and Hard. This classification is based on the complexity of information retrieval and the depth of scientific reasoning required.

\noindent \textbf{Easy (51.4\%): Direct Recognition and Single-Step Application.} 
This category encompasses tasks that rely on direct visual recognition or single-step knowledge retrieval. These questions test whether a model possesses the fundamental ``vocabulary'' of polymer science and can apply basic principles to clear, unambiguous visual stimuli. A representative example is the Foundational Knowledge Application module (as shown in Figure~\ref{fig:case}, Case 1). Here, the reasoning process is straightforward: once the model comprehends the fundamental definition of a contact angle, it can directly recognize the visual feature (i.e., the magnitude of the angle) and perform the ranking task immediately without requiring complex intermediate logic or derivation.

\noindent \textbf{Medium (30.3\%): Multi-Step Reasoning and Cross-Modal Alignment.} 
The medium-tier tasks elevate the challenge by requiring multi-step reasoning, cross-modal alignment, and specific domain application. Unlike the ``Easy'' tasks, the answer cannot be retrieved directly; it must be deduced through a chain of thought that bridges visual cues with domain-specific knowledge. This level is best exemplified by the Experiment Mechanism Reasoning module (see Figure~\ref{fig:1}). In these scenarios, the model must parse a professional reaction diagram, align the text annotations with chemical structures, and reason through the reaction pathway to infer intermediate steps or catalytic cycles.

\noindent \textbf{Hard (18.3\%): Holistic Synthesis and Multidisciplinary Analysis.} 
These ``Challenge Problems'' represent the bottleneck of current AI systems, requiring expert-level synthesis to interpret highly unstructured, noisy, or abstract data within complex contexts. This category primarily includes Raw Data Extraction, Performance and Application Exploration, and notably, Lab Safety Analysis. For instance, in safety analysis tasks, the model must navigate cluttered, real-world laboratory environments. It requires a holistic, multidisciplinary understanding (integrating chemical properties, physical principles such as airflow dynamics, and safety protocols) to identify context-dependent hazards that are not explicitly labeled. Similarly, interpreting raw spectra requires the model to rigorously filter signal from noise, handling severe data ambiguity that typically stumps generalist models.


\begin{table*}[htbp]
\centering
\caption{Performance comparison of different models across three difficulty levels (\textbf{Hard}, \textbf{Medium}, \textbf{Easy}) and the \textbf{Overall} weighted average. \textbf{P}, \textbf{R}, and \textbf{F1} represent Precision, Recall, and F1-score, respectively. The best results are in \textbf{bold} and the second best are \underline{underlined}.}
\label{tab:difficulty_results}
\resizebox{\textwidth}{!}{%
\begin{tabular}{lrrrrrrrrrrrr}
\toprule
& \multicolumn{3}{c}{\textbf{Hard}} & \multicolumn{3}{c}{\textbf{Medium}} & \multicolumn{3}{c}{\textbf{Easy}} & \multicolumn{3}{c}{\textbf{Overall}} \\
\cmidrule(lr){2-4} \cmidrule(lr){5-7} \cmidrule(lr){8-10} \cmidrule(lr){11-13}
\textbf{Model} & \textbf{P} & \textbf{R} & \textbf{F1} & \textbf{P} & \textbf{R} & \textbf{F1} & \textbf{P} & \textbf{R} & \textbf{F1} & \textbf{P} & \textbf{R} & \textbf{F1} \\
\midrule
\multicolumn{13}{c}{\textit{\textbf{Closed-source Models}}} \\
\midrule
\textbf{O3} & 0.458 & \underline{0.503} & \textbf{0.430} & 0.544 & 0.672 & \textbf{0.561} & 0.621 & 0.853 & \textbf{0.686} & 0.567 & 0.733 & \textbf{0.601} \\
\textbf{GPT-5} & 0.432 & \textbf{0.546} & \underline{0.419} & 0.483 & \underline{0.692} & \underline{0.520} & 0.590 & \underline{0.876} & \underline{0.669} & 0.529 & \underline{0.760} & \underline{0.578} \\
\textbf{Claude-Sonnet-4.5-Thinking} & 0.392 & 0.432 & 0.365 & 0.480 & 0.638 & 0.513 & 0.559 & 0.820 & 0.632 & 0.503 & 0.692 & 0.546 \\
\textbf{Gemini-2.5-Pro} & 0.416 & 0.455 & 0.378 & 0.465 & 0.679 & 0.512 & 0.519 & 0.853 & 0.608 & 0.483 & 0.726 & 0.536 \\
\textbf{Grok-4} & 0.464 & 0.341 & 0.338 & 0.567 & 0.524 & 0.495 & \underline{0.669} & 0.676 & 0.625 & \underline{0.600} & 0.568 & 0.533 \\
\textbf{Gemini-2.5-Flash} & 0.330 & 0.341 & 0.289 & 0.425 & \textbf{0.716} & 0.504 & 0.464 & \textbf{0.981} & 0.598 & 0.427 & \textbf{0.783} & 0.512 \\
\textbf{Gemini-2.0-Flash-Thinking} & \textbf{0.477} & 0.286 & 0.303 & \textbf{0.598} & 0.401 & 0.423 & \textbf{0.696} & 0.540 & 0.555 & \textbf{0.625} & 0.450 & 0.468 \\
\textbf{GPT-4o} & \underline{0.469} & 0.180 & 0.219 & \underline{0.582} & 0.298 & 0.345 & 0.636 & 0.396 & 0.436 & 0.587 & 0.323 & 0.367 \\
\textbf{GPT-4o-mini} & 0.412 & 0.175 & 0.207 & 0.497 & 0.244 & 0.283 & 0.537 & 0.317 & 0.343 & 0.501 & 0.268 & 0.299 \\
\midrule
\multicolumn{13}{c}{\textit{\textbf{Open-source Models}}} \\
\midrule
\textbf{Qwen3-VL-235B-A22B-Thinking} & \underline{0.443} & \underline{0.390} & \textbf{0.359} & \underline{0.527} & \textbf{0.558} & \textbf{0.498} & \textbf{0.620} & \textbf{0.729} & \textbf{0.628} & \underline{0.558} & \textbf{0.615} & \textbf{0.538} \\
\textbf{Qwen3-VL-32B-Thinking} & 0.408 & \textbf{0.405} & \underline{0.355} & 0.509 & \underline{0.550} & \underline{0.490} & 0.577 & \underline{0.723} & \underline{0.608} & 0.525 & \underline{0.612} & \underline{0.525} \\
\textbf{DeepSeek-R1} & 0.366 & 0.393 & 0.339 & 0.464 & 0.548 & 0.462 & 0.503 & 0.705 & 0.549 & 0.466 & 0.600 & 0.484 \\
\textbf{Intern-S1} & 0.429 & 0.258 & 0.262 & 0.531 & 0.433 & 0.400 & 0.599 & 0.573 & 0.499 & 0.548 & 0.473 & 0.427 \\
\textbf{Qwen2.5-VL-72B-Instruct} & \textbf{0.436} & 0.190 & 0.222 & \textbf{0.564} & 0.270 & 0.318 & \underline{0.600} & 0.349 & 0.390 & \textbf{0.559} & 0.295 & 0.337 \\
\textbf{DeepSeek-VL2} & 0.291 & 0.077 & 0.096 & 0.329 & 0.095 & 0.117 & 0.384 & 0.127 & 0.153 & 0.350 & 0.108 & 0.132 \\
\bottomrule
\end{tabular}%
}
\end{table*}

\subsection{Data Distribution and Difficulty Analysis}
\label{sec:data_dist}

We analyze the effectiveness of the dataset distribution through the lens of model performance reported in Table~\ref{tab:difficulty_results}. The experimental results empirically validate the distinctiveness of the three difficulty tiers.

\begin{figure*}[t]
    \centering
    \includegraphics[width=0.95\linewidth]{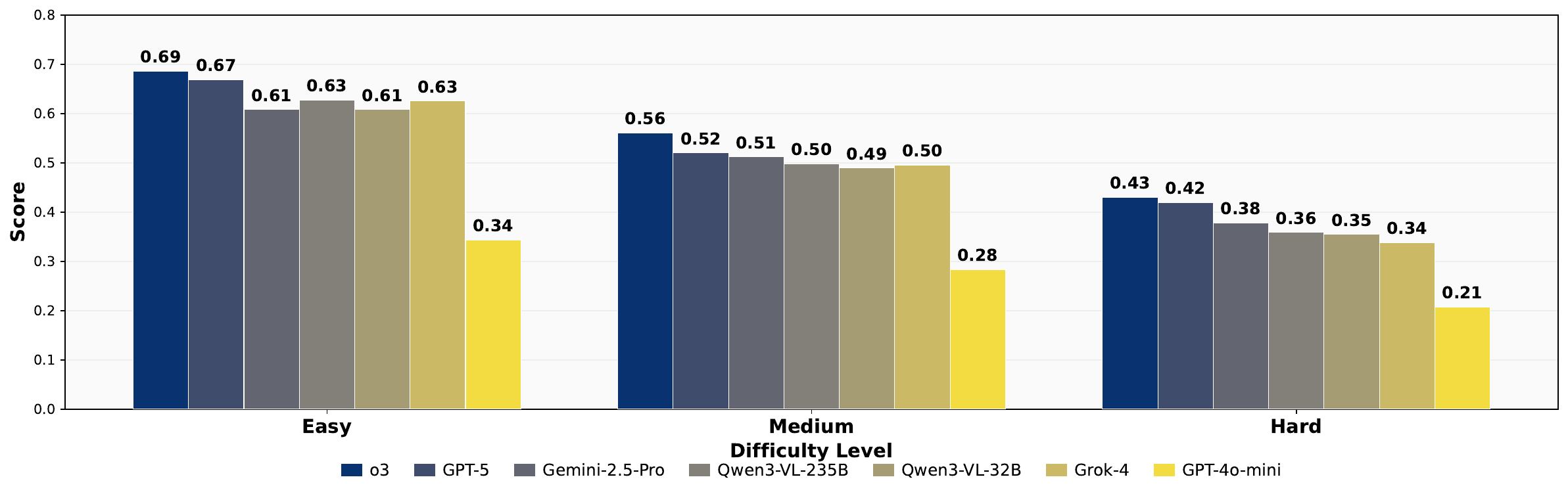} 
    \vspace{-2mm}
    \caption{\textbf{Performance comparison across difficulty levels.} F1-scores for representative models are reported across Easy, Medium, and Hard subsets. A consistent performance degradation is observed as task complexity increases, with the "Hard" subset effectively differentiating robust state-of-the-art models (e.g., o3) from lightweight counterparts (e.g., GPT-4o-mini) which experience steeper declines.}
    \label{fig:difficulty_breakdown}
\end{figure*}

\textbf{Validation of Difficulty Stratification.} 
The results in Table~\ref{tab:difficulty_results} demonstrate a consistent monotonic degradation in performance across all 15 evaluated models as task difficulty increases, confirming that the stratification criteria effectively map to model capabilities. For instance, the state-of-the-art model O3 exhibits a clear step-wise decline in F1-score, dropping from 0.686 on the Easy subset to 0.561 on Medium, and finally to 0.430 on Hard. This trend indicates that the "Hard" subset successfully isolates complex reasoning and holistic synthesis from simpler perceptual tasks(As shown in Figure~\ref{fig:difficulty_breakdown}). Furthermore, the "Hard" category acts as a critical discriminator between closed-source and open-source models. While top-tier proprietary models maintain viable performance on these complex tasks, many open-weights models experience a severe capability collapse (e.g., DeepSeek-VL2 drops to an F1-score of 0.096), highlighting that the benchmark possesses a high ceiling necessary for evaluating future agents with strong reasoning abilities.


\textbf{Disparity Between Recall and Precision.} 
A critical behavioral trend observed in Table~\ref{tab:difficulty_results} is the systematic divergence between Recall ($R$) and Precision ($P$). This phenomenon is predominantly manifested in the "Easy" tier, where models consistently achieve disproportionately high Recall scores. For instance, Gemini-2.5-Flash achieves a near-perfect Recall of 0.981, yet its Precision remains remarkably low at 0.464. This gap suggests a prevalence of "defensive verbosity" in model responses. When addressing domain-specific scientific queries, generalist MLLMs tend to generate extensive, exhaustive explanations to ensure all potential key points are covered. While this strategy successfully retrieves the correct information in simpler contexts (thereby boosting Recall), it inevitably introduces a substantial amount of redundant context or subtle factual hallucinations, which severely penalizes Precision. In practical polymer science workflows, this behavior is suboptimal, as it forces researchers to manually filter relevant insights from noisy outputs.

\noindent Conversely, the performance dynamics shift dramatically within the "Hard" difficulty tier, where we observe a precipitous decline in Recall alongside low Precision. As shown in Table~\ref{tab:difficulty_results}, even capable models like GPT-4o see their Recall drop to 0.180 on Hard tasks, while open-source models such as DeepSeek-VL2 fall to a Recall of just 0.077. This "Recall Collapse" indicates that the strategy of verbose generation becomes ineffective when models face complex, multidisciplinary challenges (e.g., holistic synthesis or mechanism reasoning). In these scenarios, models fail to identify the core scientific principles entirely, resulting in the omission of essential Key Points rather than merely burying them in noise. This distinction strongly validates the robustness of our expert-defined difficulty stratification: the "Hard" subset successfully exposes the absolute reasoning boundaries of current MLLMs, confirming that future improvements must focus on deepening genuine scientific reasoning capabilities rather than simply optimizing for conversational comprehensiveness.

\section{Implementation and Evaluation Details}


This section outlines the comprehensive protocols ensuring the reproducibility and rigor of the \textit{PolyReal} benchmark. We detail the standardized inference setup and the automated evaluation pipeline, emphasizing how specific prompting strategies were employed to elicit deep reasoning and quantify scientific accuracy.

\definecolor{cvprblue}{RGB}{13, 71, 161}
\definecolor{brickred}{RGB}{192, 57, 43}    

\begin{figure}[h!]
\centering
\begin{tcolorbox}[
    colback=white, 
    colframe=cvprblue, 
    title={\textbf{Prompt for PolyReal Inference}}, 
    width=0.98\linewidth,
    arc=2pt,                
    boxrule=0.8pt,          
    left=6pt, right=6pt, top=6pt, bottom=6pt, 
    fonttitle=\bfseries,
    fontupper=\small\sffamily 
]
\textbf{Role:} You are a polymer science expert. Your task is to provide a clear and accurate answer.

\vspace{0.5em} 
\textbf{Process:}
\begin{itemize}[leftmargin=1.5em, itemsep=2pt, topsep=2pt] 
    \item \textbf{Internal Reasoning} (inside the \textcolor{cvprblue}{\texttt{<think>}} tag): \\
    Lay out your step-by-step reasoning process here.
    
    \item \textbf{Final Synthesized Answer} (inside the \textcolor{cvprblue}{\texttt{<answer>}} tag): \\
    After your reasoning, place the well-organized, clear, accurate, and concise answer within the tag. This answer must be a standalone, concise, and professional explanation that directly addresses the user's question. Do not simply repeat the reasoning process. You should distill the key conclusions from your thinking process to form a polished response.
\end{itemize}

\vspace{0.5em}
\textbf{Constraint:} Please ensure your final response includes both the complete \textcolor{cvprblue}{\texttt{<think>}} and \textcolor{cvprblue}{\texttt{<answer>}} sections.
\end{tcolorbox}
\caption{The standardized system prompt used for zero-shot inference across all MLLMs, enforcing a structured Chain-of-Thought reasoning process.}
\label{fig:inference_prompt}
\end{figure}

\subsection{Experimental Setup}
\label{sec:exp_setup}
To evaluate the depth of scientific reasoning, we moved beyond standard question-answering setups by enforcing a strict Chain-of-Thought (CoT) protocol. As illustrated in \textbf{Figure~\ref{fig:inference_prompt}}, we designed a unified system prompt that mandates a dual-stage output format. 

\textbf{Enforcing Explicit Reasoning.} 
The prompt explicitly instructs all MLLMs to separate their internal thought process (enclosed in \texttt{<think>} tags) from their final conclusion (enclosed in \texttt{<answer>} tags). This structural constraint serves two purposes:
\begin{enumerate}
    \item \textbf{Qualitative Diagnosis:} It allows us to inspect the model's intermediate logic, identifying whether an error stems from a hallucinated visual feature or a flaw in reasoning.
    \item \textbf{Persona Alignment:} By defining the role of a "polymer science expert," the prompt conditions the model to adopt a professional tone and prioritize scientific rigor over conversational casualness.
\end{enumerate}

\begin{figure}[htbp]
\centering
\begin{tcolorbox}[
    enhanced,
    colback=white, 
    colframe=cvprblue, 
    title={\textbf{Automated Evaluator Prompt: Recall}}, 
    width=0.98\linewidth,
    arc=2pt, boxrule=0.8pt,
    left=6pt, right=6pt, top=6pt, bottom=6pt,
    fonttitle=\bfseries,
    fontupper=\scriptsize\sffamily 
]
\textbf{Role:} You are an exceptionally strict, meticulous, and critical grader specializing in polymer science. Your task is to evaluate a "Model's Answer" based on a list of "Key Scoring Points" and a "Ground Truth Answer".

\vspace{0.4em}
\textbf{Evaluation Task:} You must evaluate \textit{each} "Key Scoring Point" \textit{sequentially} based on the following two dimensions:

\vspace{0.3em}
\textbf{1. Completeness (\textcolor{cvprblue}{met} - Binary Score):}
\begin{itemize}[leftmargin=1.5em, nosep]
    \item Does the "Model's Answer" clearly and unambiguously cover the core concept of this "Key Scoring Point"? This is a strict binary (0 or 1) check.
    \item \textbf{\textcolor{cvprblue}{1 (Met):}} The point is clearly and directly addressed, and its explanation has no critical information missing compared to the relevant explanation of this point in the "Ground Truth Answer".
    \item \textbf{\textcolor{brickred}{0 (Not Met):}} The point is missing, glossed over, vaguely implied, or has significant information omissions compared to the "Ground Truth Answer".
\end{itemize}

\vspace{0.3em}
\textbf{2. Professional Quality (\textcolor{cvprblue}{quality\_score} - Float 0.0 to 1.0):}
\begin{itemize}[leftmargin=1.5em, nosep]
    \item \textbf{If \textcolor{cvprblue}{met} is 0}, this score \textbf{MUST} be \textbf{\textcolor{brickred}{0.0}}.
    \item \textbf{If \textcolor{cvprblue}{met} is 1}, you must then grade \textit{how well} the point was covered:
    \begin{itemize}[leftmargin=1em, nosep]
        \item \textbf{\textcolor{cvprblue}{1.0 (Perfect):}} The explanation is impeccable in fact, depth, and accuracy. The terminology is professional, the logic is rigorous.
        \item \textbf{\textcolor{orange!80!black}{0.5 (Average):}} The point is covered, but the explanation is superficial, imprecise, logically flawed, uses casual language, or is vague.
        \item \textbf{\textcolor{brickred}{0.1 (Poor):}} The point is mentioned, but its explanation contains severe factual errors or logical fallacies.
    \end{itemize}
\end{itemize}

\vspace{0.5em}
\textbf{Your Evaluation (in JSON format only):} 
Output a SINGLE JSON object containing exactly three keys:
\begin{itemize}[leftmargin=1.5em, nosep]
    \item \textbf{\textcolor{cvprblue}{"met"}}: [List of integers 0 or 1]
    \item \textbf{\textcolor{cvprblue}{"quality\_score"}}: [List of floats between 0.0 and 1.0]
    \item \textbf{\textcolor{cvprblue}{"reasoning"}}: "A step-by-step, critical explanation. If \textcolor{brickred}{'Not Met'}, clearly specify the failure."
\end{itemize}
\end{tcolorbox}
\caption{The full prompt template for the Precision Evaluator. We explicitly categorize distinct types of noise (highlighted in \textcolor{brickred}{red}) to rigorously penalize hallucinations and ensuring the Precision metric reflects true information density.}
\label{fig:prompt_recall_full}
\end{figure}

\begin{figure}[htbp]
\centering
\begin{tcolorbox}[
    enhanced,
    colback=white, 
    colframe=cvprblue, 
    title={\textbf{Automated Evaluator Prompt: Precision}}, 
    width=0.98\linewidth,
    arc=2pt, boxrule=0.8pt,
    left=6pt, right=6pt, top=6pt, bottom=6pt,
    fonttitle=\bfseries,
    fontupper=\scriptsize\sffamily
]
\textbf{Role:} You are a rigorous, fair, and professional Benchmark Evaluator. Your task is to calculate the "Precision" of a "Model's Answer" and verify its coverage of the "Key Scoring Points."

\vspace{0.4em}
\textbf{Core Calculation Formula:} $\text{Precision} = \frac{\text{Total TP Count}}{\text{Total TP Count} + \text{Total FP Count}}$

\vspace{0.4em}
\textbf{Evaluation Criteria:} You must strictly adhere to the following definitions to count TP and FP:

\begin{itemize}[leftmargin=1.5em, itemsep=2pt]
    \item \textbf{\textcolor{cvprblue}{TP (True Positive):}} 
    A specific unit of information (a phrase or sentence) within the "Model's Answer" that clearly and directly corresponds to one of the points in the "Key Scoring Points (Keywords)" list.

    \item \textbf{\textcolor{brickred}{FP (False Positive):}} 
    A specific unit of information within the "Model's Answer" that falls into any of the following categories:
    \begin{itemize}[leftmargin=1em, nosep]
        \item \textbf{[\textcolor{brickred}{FP-Irrelevant}]:} The information is correct but irrelevant to the current question.
        \item \textbf{[\textcolor{brickred}{FP-Incorrect}]:} A factual error or hallucination.
        \item \textbf{[\textcolor{brickred}{FP-Redundant}]:} A verbose, repetitive restatement of the same point already counted as a TP.
        \item \textbf{[\textcolor{brickred}{FP-Filler}]:} "Empty" phrases with no informational value (e.g., "This is a good question," "In conclusion").
    \end{itemize}
\end{itemize}

\vspace{0.4em}
\textbf{Note:} Points from the Key Scoring Points list that are missed (False Negatives) do not participate in the Precision calculation.

\vspace{0.4em}
\textbf{Your Evaluation (in JSON format only):}
Your ONLY task is to identify all TP and FP information units. Output a single JSON object:
\begin{itemize}[leftmargin=1.5em, nosep]
    \item \textbf{\textcolor{cvprblue}{"tp\_string"}}: "(Found first TP unit)..."
    \item \textbf{\textcolor{cvprblue}{"fp\_string"}}: "(Found first FP unit - [FP-Type])..."
\end{itemize}
\end{tcolorbox}
\caption{Full prompt for the Precision Evaluator. The explicit categorization of False Positives (highlighted in red) ensures that verbose or hallucinatory content is accurately penalized.}
\label{fig:prompt_precision_full}
\end{figure}

\subsection{Automated Evaluation Pipeline}
\label{sec:eval_prompt}

Given the open-ended nature of the tasks, we implemented a fine-grained "LLM-as-a-Judge" framework. Unlike simple keyword matching, our pipeline decouples the assessment of Completeness (Recall) from Correctness (Precision) using two distinct evaluator prompts.

\textbf{Recall and Quality Assessment.} 
As detailed in \textbf{Figure~\ref{fig:prompt_recall_full}}, the Recall Evaluator acts as a "strict grader." It verifies whether the model's response covers the expert-defined Key Scoring Points. Crucially, this prompt goes beyond a binary check; it includes a \texttt{quality\_score} (0.0–1.0) to penalize correct but superficial explanations. This ensures that models are rewarded not just for mentioning a keyword, but for demonstrating a deep understanding of the underlying concept.

\textbf{Precision and Noise Filtering.} 
To address the "defensive verbosity" issue (where models generate excessive text to mask uncertainty), we designed a specialized Precision Evaluator shown in \textbf{Figure~\ref{fig:prompt_precision_full}}. This prompt explicitly categorizes "False Positive" (FP) information into four distinct types: \textit{Irrelevant}, \textit{Incorrect}, \textit{Redundant}, and \textit{Filler}. By strictly penalizing these noise categories, we obtain a precise measure of the signal-to-noise ratio, distinguishing models that truly "know" the answer from those that simply "guess and expand."

\subsection{Human Annotation Guidelines}
\label{sec:human_annotation}

Complementing the data sourcing strategy detailed in Section~\ref{sec:data_collection} of the main paper, this section outlines the rigorous annotation protocols followed by our expert team. To ensure the benchmark's reliability, we established a standardized workflow for converting raw scientific data into structured evaluable tasks.

\textbf{Expert-Driven Curation.} 
The annotation process was conducted exclusively by domain experts with deep specialization in polymer science, encompassing sub-fields such as macromolecular chemistry, spectroscopy, and materials engineering. Unlike general-purpose benchmarks, no non-expert annotators were involved. This ensured that the interpretation of complex instrument data (e.g., assigning NMR signals or analyzing reaction mechanisms) adhered to professional scientific standards.

\textbf{Guideline for ``Key Points" Extraction.} 
To support the fine-grained evaluation metrics described in Section~\ref{sec:eval_prompt}, experts were required to structure the Ground Truth Answer not as a monolithic block of text, but as a sequence of verifiable facts.
\begin{itemize}[leftmargin=*, nosep]
    \item \textbf{Atomic Decomposition:} Annotators decomposed complex reasoning chains into independent "Key Scoring Points." For example, in a spectral analysis task, identifying a specific peak wavenumber constitutes one point, while correctly assigning it to a functional group constitutes another.
    \item \textbf{Fact Verification:} Each key point was required to be explicitly supported by visual evidence in the provided data or by established chemical principles, strictly prohibiting ambiguous or subjective statements.
\end{itemize}

\textbf{Cross-Validation and Quality Assurance.} 
We implemented a "Peer-Review" validation protocol mirroring academic publication standards. After the initial authoring, each sample underwent a blind review by a second expert who attempted to solve the problem solely based on the visual input. Samples were flagged for revision if the second expert identified factual inaccuracies, ambiguous visual cues, or if the "Key Scoring Points" were insufficient to cover a complete scientific answer. Only samples achieving consensus were retained in the final dataset.

\section{Additional Qualitative Analysis}
\begin{figure}[htbp]
\centering
\begin{tcolorbox}[
    enhanced,
    colback=white, 
    colframe=cvprblue, 
    title={\textbf{Lab Safety Analysis}}, 
    width=0.98\linewidth,
    arc=2pt, boxrule=0.8pt,
    left=6pt, right=6pt, top=6pt, bottom=6pt,
    fonttitle=\bfseries,
    fontupper=\small\sffamily
]
    \centering
    \includegraphics[width=0.95\linewidth]{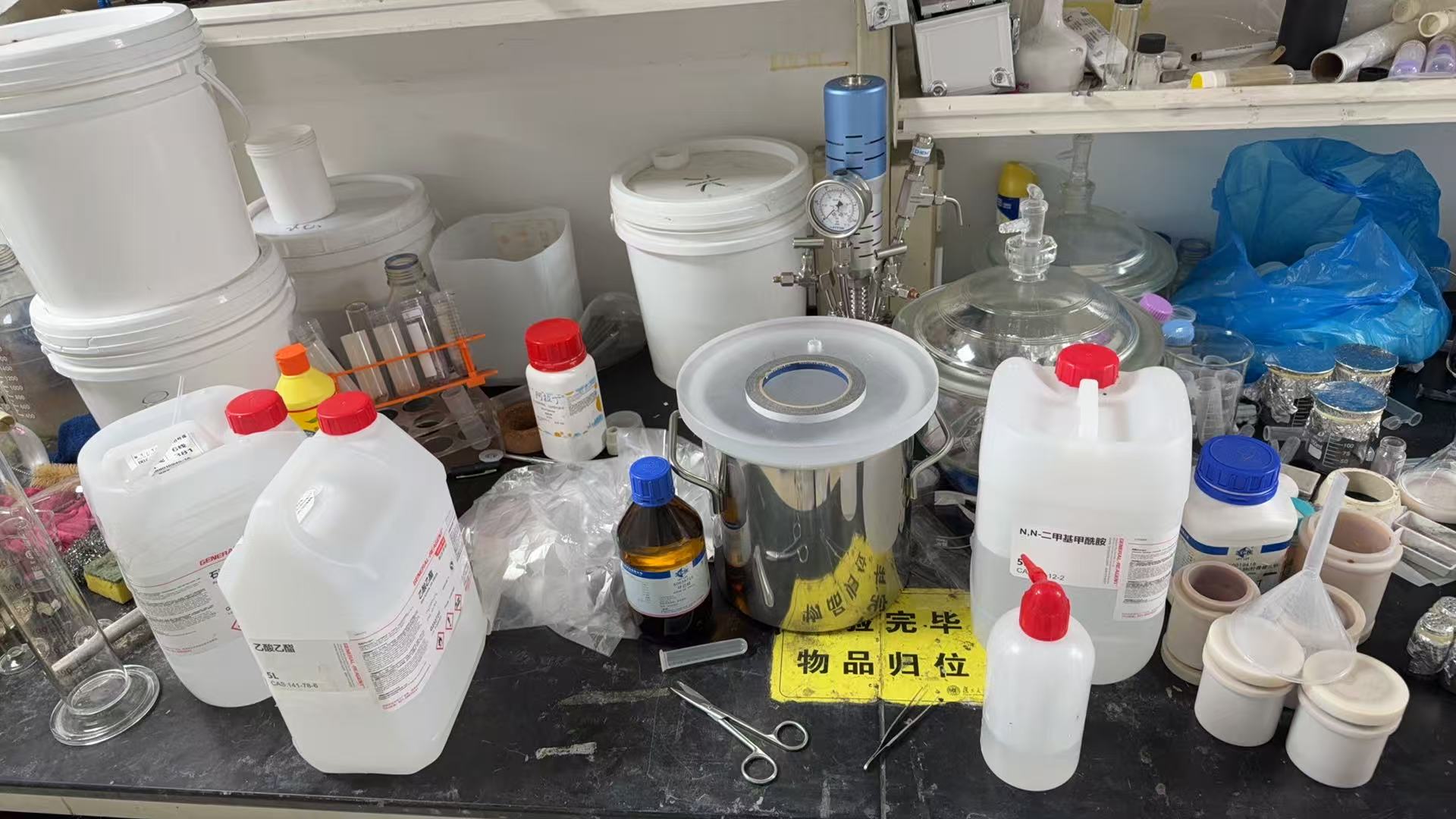}
    \vspace{0.5em}

    \raggedright
    \textbf{Question:} \\
    You are operating on a 'polymerization completion processing' experimental bench... This bench contains large quantities of solvents commonly used in polymer synthesis... containing \textcolor{brickred}{severe safety hazards}... Please identify and analyze their causes.

    \vspace{0.5em}
    \textbf{Answer:} \\
    Large-capacity (several liters) '\textcolor{cvprblue}{ethyl acetate}' (\textcolor{brickred}{highly flammable}) and '\textcolor{cvprblue}{N,N-dimethylformamide}' (\textcolor{cvprblue}{DMF}, \textcolor{brickred}{flammable} and possessing \textcolor{brickred}{reproductive toxicity/hepatotoxicity}) are directly piled on the workbench... \textcolor{cvprblue}{ethyl acetate} vapors are extremely susceptible to \textcolor{brickred}{ignition}... meanwhile, the \textcolor{brickred}{toxic vapors} of \textcolor{cvprblue}{DMF} will \textcolor{brickred}{continuously volatilize}... Potential risk of \textcolor{brickred}{mixed storage of incompatible chemicals}... there is an extremely high risk of \textcolor{brickred}{incompatible chemicals} (such as \textcolor{cvprblue}{strong oxidizers} and \textcolor{cvprblue}{flammable solvents}) being mistakenly placed together, which could lead to \textcolor{brickred}{exothermic reactions}, \textcolor{brickred}{fires}, or the \textcolor{brickred}{release of toxic gases}.

    \vspace{0.5em}
    \textbf{Key Points:}
    \begin{itemize}[leftmargin=1.5em, nosep]
        \item Accurately identify large-capacity flammable solvent '\textcolor{cvprblue}{ethyl acetate}' on the bench.
        \item Accurately identify large-capacity toxic solvent '\textcolor{cvprblue}{N,N-dimethylformamide (DMF)}'.
        \item Clearly point out the core mistake: large-capacity solvents should be stored in safety cabinets, not on the bench.
        \item Analyze the \textcolor{brickred}{severe fire and explosion risk} of flammable solvent (\textcolor{cvprblue}{ethyl acetate}) exposed.
        \item Analyze the \textcolor{brickred}{acute or chronic poisoning health risk} of toxic solvent (\textcolor{cvprblue}{DMF}) exposed.
        \item ....
    \end{itemize}
\end{tcolorbox}
\caption{Qualitative analysis of a failure case. Specific \textcolor{cvprblue}{chemical substances} are highlighted in blue, and identified \textcolor{brickred}{safety hazards} are highlighted in red.}
\label{fig:safety_case_study}
\end{figure}

A recurring failure mode observed in \textit{PolyReal} is "Scientific Hallucination," where models generate plausible-sounding but factually non-existent evidence. Unlike general-domain hallucinations, these errors in polymer science typically stem from a conflict between the model's internal chemical priors and the specific visual data provided. We categorize these into two primary types:

\subsection{Prior-Dominated Hallucination}
This type occurs when the model prioritizes textual hints over visual evidence, leading to a "force-fit" analysis.
\begin{itemize}
    \item \textbf{Mechanism:} When provided with a strong textual cue like a molecular formula (e.g., "C$_2$H$_6$OS" in Figure~\ref{fig:case}, Case 2), the model acts on statistical intuition rather than visual perception. It retrieves a plausible isomer from its internal knowledge base and attempts to rationalize the visual spectrum to match this pre-conceived structure.
    
    \item \textbf{Manifestation:} In the referenced NMR analysis case, the model hallucinated the existence of specific spectral peaks to justify its structural prediction. Despite the visual data showing a simpler signal pattern, the model fabricated descriptions of non-existent signals to maintain consistency with its initial hypothesis. This demonstrates a critical weakness in cross-modal alignment, where the model effectively "sees what it expects to see" rather than what is actually present.
\end{itemize}

\subsection{Logical Over-Inference}
This type arises when models bridge visual gaps using semantic probability, particularly in high-stakes scenarios like Lab Safety.

\noindent\textbf{Case Analysis}
To concretely illustrate this, we analyze an excerpt from the SOTA model O3's response to the Safety task (refer to the visual input in Figure~\ref{fig:safety_case_study}). While O3 correctly identified the major hazards, its supporting evidence contained significant fabrications driven by a ``Safety Alignment Bias"

\noindent\textbf{O3 Response:}
\begin{quote}
\small
\textit{"Major catastrophic-level safety hazards... \textcolor{cvprblue}{Several 5 L jerrycans of ethyl acetate are kept open or loosely capped} on the open bench... Ignition sources are present (motor on the pressure reactor, \textcolor{brickred}{possible heating mantle, metal tools, static discharge from plastic})... If over-pressurised, mechanically damaged or \textcolor{brickred}{its relief valve vents}, it can violently eject hot monomer/solvent aerosol."}
\end{quote}

\noindent This excerpt reveals how the model "over-interprets" visual cues to align with a cautious safety narrative:

\begin{itemize}
    \item \textbf{Visual Distortion (\textcolor{cvprblue}{Blue}):} The model correctly identifies the object ("Ethyl Acetate") but distorts its attributes to support a "Worst-Case Scenario." It exaggerates quantity ($1 \to \textit{Several}$) and misinterprets the state ($Closed \to \textit{Kept Open}$). By hallucinating that the containers are open, the model constructs a more compelling justification for the "Fire Risk," prioritizing a conservative safety warning over factual visual reporting.
    
    \item \textbf{Associative Hallucination (\textcolor{brickred}{Red}):} The model invents objects like a "\textit{heating mantle}" or "\textit{relief valve}" which are visually absent. These hallucinations stem from \textbf{Probabilistic Semantic Association}: since "Reactors" co-occur with "Heating Mantles" in chemical safety literature, the model infers their presence to complete a logical "Ignition Source" chain.
\end{itemize}

\textbf{Impact.} 
This pattern poses a subtle risk: \textbf{Correct Conclusion, Fabricated Evidence.} While the safety warning is valid, the user might be misled to search for non-existent hazards (e.g., a heating mantle) while overlooking real ones. This necessitates the rigorous \textit{Key Scoring Point} verification protocol used in \textit{PolyReal} to penalize invented data.

\section{Small-Model Results}
To complement the main results, we additionally evaluate smaller models in the 2B--13B range. As shown in Table~\ref{tab:model_performance}, the results exhibit a clear scaling trend: performance improves substantially with model size. Moreover, at comparable scales, reasoning-oriented models generally outperform their standard counterparts, suggesting that explicit reasoning remains beneficial in the small-model regime.

\begin{table}[h]
\centering
\caption{Performance of additional small models on PolyReal.}
\label{tab:model_performance}
\resizebox{\linewidth}{!}{%
\begin{tabular}{l|c|c|c}
\toprule
\textbf{Model} & \textbf{Precision (Avg.)} & \textbf{Recall (Avg.)} & \textbf{F1 (Avg.)} \\ 
\midrule
LLaVA-v1.6-Vicuna-13B & 0.110 & 0.058 & 0.053 \\
Qwen3-VL-2B-Thinking & 0.213 & 0.196 & 0.166 \\
InternVL3\_5-8B & 0.415 & 0.366 & 0.337 \\
Qwen3-VL-8B-Instruct & 0.543 & 0.345 & 0.366 \\ 
Qwen3-VL-8B-Thinking & 0.442 & 0.442 & 0.395 \\
\bottomrule
\end{tabular}%
}
\end{table}


\end{document}